\documentclass{article}

\usepackage[final]{neurips_2020}

\usepackage{natbib}
\usepackage[utf8]{inputenc} 
\usepackage[T1]{fontenc}    
\usepackage{hyperref}       
\usepackage{url}            
\usepackage{booktabs}       
\usepackage{amsfonts}       
\usepackage{nicefrac}       
\usepackage{microtype}      
\usepackage{graphicx}
\graphicspath{ {./} }

\title{A Novel Methodology For Crowdsourcing AI Models in an Enterprise}

\author{%
  Parthasarathy Suryanarayanan\\
  IBM Research\\
  \texttt{psuryan@us.ibm.com} \\
  \And
  Sundar Saranathan \\
  IBM Research\\
  \texttt{ssaranathan@us.ibm.com} \\
  \And
  Shilpa Mahatma \\
  IBM Research\\
  \texttt{mahatma@us.ibm.com} \\
   \AND
  Divya Pathak \\
  IBM Research\\
  \texttt{drpathak@us.ibm.com} \\
}

\begin{document}

\maketitle

\section{Introduction}

From finance to healthcare, many businesses are turning to artificial intelligence (AI) to grow and optimize their business operations. However, there is a shortage of AI expertise in the market~\citep{demand}. At the same time, AI technologies continue to evolve rapidly~\citep{perrault2019artificial}. Industries are increasingly relying on crowdsourcing in order to meet their AI needs~\citep{Vaughan2018crowdsourcing}. 

Competitions are a popular method for crowdsourcing AI models. In order to develop product features, major technology firms such as Microsoft~\citep{malwareChallenge}, Google~\citep{youtube8mChallenge} and Facebook~\citep{dolhansky2019deepfake} have resorted to AI competitions that often draw a vast number of participants. These competitions are usually hosted in platforms such as Kaggle, Codalab~\citep{codalab, liang2015codalab}, Eval AI~\citep{yadav2019evalai}, etc. Often these platforms are tightly coupled with a proprietary cloud infrastructure. These platforms cannot be used by organizations with stringent data regulations that require on-premise data retention. In addition, the terms and conditions governing the use of such platforms may not be amiable to the organization seeking to host a competition. 

In this work, we propose a novel method and system that any organization can easily adopt to host AI competitions. The system allows them to automatically harvest and evaluate the submitted models against in-house proprietary data and also to incorporate them as reusable services in a product.

\section{Proposed solution}

\begin{figure}[h]
\centering
\includegraphics[width=0.8\textwidth]{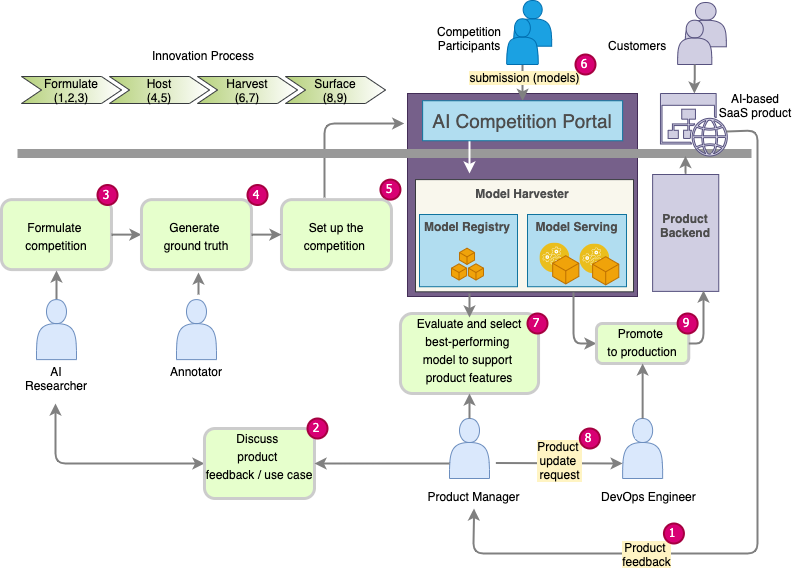}
\caption{Sequence of steps in the proposed method for leveraging crowdsourcing of AI models.}
    \label{fig:arch}
\end{figure}

In order to efficiently crowdsource AI models, our proposed approach consists of three key ideas.
\begin{enumerate}
    \item A public-facing \textbf{AI Competition Portal} operated by the organization, where key AI needs of its products are presented as competitions in the portal. It is important that the portal is hosted in an environment that is completely managed by the organization. 
    \item Each of these competitions is set up using a \textbf{common source code template} such that the participant's submission consists of a model and code to invoke the model using a standard interface. This ensures interoperability, security and the ability to tune the submission source code.
    \item A \textbf{Model Harvester} that converts each submitted model into a runnable microservice and also provides a dashboard of all available models. 
\end{enumerate}
This is depicted in the Figure~\ref{fig:arch}. In the rest of this section, each of these ideas is explained.

\textbf{AI Competition Portal}: Large organizations have many products serving different markets. The portal with well-defined competitions across product boundaries offers a centralized repository that maps and aligns AI research activities with the business needs of the organization. Organizational AI needs are formulated as competitions in the portal, inviting participation from outside researchers and data scientists. Once the competition is formulated, it is hosted in the portal with a call for public participation. External participation is encouraged through monetary rewards. Activities 1 through 5 illustrate this workflow. Each competition is set up using a code template that the participant teams must use for developing their model training and inference routines. Participant teams submit models together with the code and not just predictions. The portal can be implemented using any opensource framework such as EvalAI~\citep{yadav2019evalai} or CodaLab~\citep{codalab}. 

\textbf{Model Harvester}: Submissions are automatically pulled from the portal into a catalog of models called \textit{Model Registry}. The registry also tracks the model hyper-parameters, metrics and models binaries over time. The system includes utilities to turn each model into a micro-service after static code analysis (e.g. vulnerability scans) called \textit{Model Serving}. Product management teams can access these microservices via Swagger APIs and further evaluate them using additional proprietary datasets. An application feature can be quickly built by assembling these microservices into the production service orchestration. Activities 6 through 9 illustrate this workflow. The overall \textit{Model Harvester} system can be built on top of an AI-lifecycle management framework like MLflow~\citep{zaharia2018accelerating} or ProvDB~\citep{miao2018provdb}. 

\section{Conclusion}
The methodology described in this work facilitates rapid commercialization of crowdsourced models by providing a streamlined maturity process from problem definition to asset creation. A reference implementation of the system described is available in the form of \texttt{AI Leaderboard} ~\citep{IBMAILeaderboard}. Based on the open source EvalAI~\citep{yadav2019evalai} system, \texttt{AI Leaderboard} supports source code template based submissions. The integrated \textit{Model Harvester} subsystem is based on MLflow~\citep{zaharia2018accelerating}. Currently the system is being used for two academic AI competitions, EMNLP 2020 and ICDAR 2021. We are planning to use this platform for hosting industry challenges in healthcare domain. As part of the future work, we plan to publish the impact study from this crowdsourcing exercise.

\begin{ack}
The authors would like to acknowledge Ansu Varghese, Abhishek Malvankar, Ching-Huei Tsou, Jian Min Jiang, Jian Wang, Michele Payne and Sreeram Joopudi for their contributions.  
\end{ack}
\bibliography{ref}

\begin{thebibliography}{12}
\providecommand{\natexlab}[1]{#1}
\providecommand{\url}[1]{\texttt{#1}}
\expandafter\ifx\csname urlstyle\endcsname\relax
  \providecommand{\doi}[1]{doi: #1}\else
  \providecommand{\doi}{doi: \begingroup \urlstyle{rm}\Url}\fi

\bibitem[CodaLab(2017)]{codalab}
CodaLab.
\newblock Codalab.
\newblock \url{https://github.com/codalab/codalab-competitions}, 2017.

\bibitem[Dolhansky et~al.(2019)Dolhansky, Howes, Pflaum, Baram, and
  Ferrer]{dolhansky2019deepfake}
Brian Dolhansky, Russ Howes, Ben Pflaum, Nicole Baram, and Cristian~Canton
  Ferrer.
\newblock The deepfake detection challenge (dfdc) preview dataset.
\newblock \emph{arXiv preprint arXiv:1910.08854}, 2019.

\bibitem[Gartner(2019)]{demand}
Gartner.
\newblock {2019 CIO Survey: CIOs Have Awoken to the Importance of AI}.
\newblock \url{https://www.gartner.com/document/3897266}, 2019.

\bibitem[IBM(2020)]{IBMAILeaderboard}
IBM.
\newblock {AI} leaderboard.
\newblock \url{https://ibm.biz/ai-leaderboard}, 2020.

\bibitem[Lee et~al.(2018)Lee, Reade, Sukthankar, Toderici,
  et~al.]{youtube8mChallenge}
Joonseok Lee, Walter Reade, Rahul Sukthankar, George Toderici, et~al.
\newblock The 2nd youtube-8m large-scale video understanding challenge.
\newblock In \emph{Proceedings of the European Conference on Computer Vision
  (ECCV)}, pages 0--0, 2018.

\bibitem[Liang and Viegas()]{liang2015codalab}
Percy Liang and Evelyne Viegas.
\newblock Codalab worksheets for reproducible, executable papers, december
  2015.
\newblock In \emph{URL: https://nips. cc/Conferences/2015/Schedule}.

\bibitem[Miao and Deshpande(2018)]{miao2018provdb}
Hui Miao and Amol Deshpande.
\newblock Provdb: Provenance-enabled lifecycle management of collaborative data
  analysis workflows.
\newblock \emph{IEEE Data Eng. Bull.}, 41\penalty0 (4):\penalty0 26--38, 2018.

\bibitem[Perrault et~al.(2019)Perrault, Shoham, Brynjolfsson, Clark,
  Etchemendy, Grosz, Lyons, Manyika, Mishra, and
  Niebles]{perrault2019artificial}
R~Perrault, Y~Shoham, E~Brynjolfsson, J~Clark, J~Etchemendy, B~Grosz, T~Lyons,
  J~Manyika, S~Mishra, and JC~Niebles.
\newblock Artificial intelligence index report 2019, 2019.

\bibitem[Ronen et~al.(2018)Ronen, Radu, Feuerstein, Yom-Tov, and
  Ahmadi]{malwareChallenge}
Royi Ronen, Marian Radu, Corina Feuerstein, Elad Yom-Tov, and Mansour Ahmadi.
\newblock Microsoft malware classification challenge.
\newblock \emph{arXiv preprint arXiv:1802.10135}, 2018.

\bibitem[Vaughan(2018)]{Vaughan2018crowdsourcing}
J.W. Vaughan.
\newblock Making better use of the crowd: How crowdsourcing can advance machine
  learning research.
\newblock \emph{Journal of Machine Learning Research}, 18:\penalty0 1--46, 05
  2018.

\bibitem[Yadav et~al.(2019)Yadav, Jain, Agrawal, Chattopadhyay, Singh, Jain,
  Singh, Lee, and Batra]{yadav2019evalai}
Deshraj Yadav, Rishabh Jain, Harsh Agrawal, Prithvijit Chattopadhyay, Taranjeet
  Singh, Akash Jain, Shiv~Baran Singh, Stefan Lee, and Dhruv Batra.
\newblock Evalai: Towards better evaluation systems for ai agents.
\newblock \emph{arXiv preprint arXiv:1902.03570}, 2019.

\bibitem[Zaharia et~al.(2018)Zaharia, Chen, Davidson, Ghodsi, Hong, Konwinski,
  Murching, Nykodym, Ogilvie, Parkhe, et~al.]{zaharia2018accelerating}
Matei Zaharia, Andrew Chen, Aaron Davidson, Ali Ghodsi, Sue~Ann Hong, Andy
  Konwinski, Siddharth Murching, Tomas Nykodym, Paul Ogilvie, Mani Parkhe,
  et~al.
\newblock Accelerating the machine learning lifecycle with mlflow.
\newblock \emph{IEEE Data Eng. Bull.}, 41\penalty0 (4):\penalty0 39--45, 2018.

\end{thebibliography}

\end{document}